\documentclass[a4paper,11pt]{article}
\usepackage[utf8]{inputenc}
\usepackage{amsmath, amssymb}
\usepackage{listings}
\usepackage{xcolor}
\usepackage{hyperref}
\usepackage{graphicx} 
\usepackage{geometry}
\usepackage{titlesec}
\usepackage{float} 
\usepackage{bold-extra}
\usepackage{url}  

\titleformat{\section}{\large\bfseries}{\thesection.}{1em}{}
\titleformat{\subsection}{\normalsize\bfseries}{\thesubsection.}{1em}{}

\lstdefinestyle{mystyle}{
    basicstyle=\ttfamily\footnotesize,
    numbers=left,
    numberstyle=\tiny\color{gray},
    stepnumber=1,
    numbersep=10pt,
    backgroundcolor=\color{white},
    showspaces=false,
    showstringspaces=false,
    showtabs=false,
    frame=single,
    rulecolor=\color{black},
    tabsize=2,
    captionpos=b,
    breaklines=true,
    breakatwhitespace=false,
    title=\lstname,
    keywordstyle=\color{blue},
    commentstyle=\color{gray},
    stringstyle=\color{red},
    escapeinside={\%*}{*)},
    morekeywords={*,...}
}

\title{\textbf{TiniScript: A Simplified Language for Educational Robotics}}
\author{
    Jesús Guzmán \\
    Universidad ESAN \\
    \texttt{13100597@ue.edu.pe}
    \and
    Gabriel Guzmán \\
    Universidad ESAN \\
    \texttt{17200646@ue.edu.pe}
}
\date{July 2024}

\begin{document}

\maketitle

\begin{abstract}
TiniScript is an intermediate programming language designed for educational robotics, aligned with STEM principles to foster integrative learning experiences. With its minimalist single-line syntax, such as F(2, 80), TiniScript simplifies robotic programming, allowing users to bypass complex code uploading processes and enabling real-time direct instruction transmission. Thanks to its preloaded interpreter, TiniScript decouples programming from hardware, significantly reducing wait times. Instructions can be sent wirelessly from any Bluetooth-enabled device —such as tablets, computers, or phones— through a serial port, making TiniScript adaptable to a variety of robots. This adaptability and ease of use not only optimize iterative and collaborative learning but also allow students to focus on creative aspects of robotics. This paper explores TiniScript’s design principles, syntax, and practical applications, highlighting its potential to make robotics programming more accessible and effective in developing critical thinking skills.
\end{abstract}

\section*{Keywords}
TiniScript, Educational Robotics, STEM Education, Programming Languages, Block-Based Programming, Microcontroller Programming.

\section{Introduction}
The constructionism theory, formulated by Seymour Papert, has been a transformative approach in education, particularly within STEM (Science, Technology, Engineering, and Mathematics) fields. This theory emphasizes learning through creation, where students engage actively by building knowledge structures through hands-on tasks and meaningful projects. One of the early milestones influenced by constructionism was the development of the Logo programming language. Logo’s simple, block-based structure enabled students to grasp fundamental programming concepts visually by manipulating blocks, establishing a foundation for educational tools that remain essential in early computer science education.

Over time, educational robotics kits, like those from LEGO Education (RCX, NXT, and EV3), have set standards for integrating physical construction with software programming. These kits demonstrate the potential of robotics in educational settings by engaging students in both mechanical assembly and logical problem-solving, thereby fostering an understanding of hardware and software as interconnected aspects of robotics. Building on this foundation, programming environments in educational robotics have largely adopted block-based interfaces. These environments simplify coding for beginners, allowing students to create programs by connecting blocks representing specific actions. Once completed, the program is uploaded to a microcontroller, enabling the robot to execute the instructions.

Despite the accessibility and popularity of block-based environments, the process of uploading and re-uploading code after each modification can interrupt the learning flow. Each update requires additional steps, from compilation to loading onto the microcontroller, often resulting in delays that limit the opportunity for real-time interaction and experimentation. This dependency on repeated code uploads also imposes technical barriers, especially for younger learners, as they need to manage these mechanical aspects in addition to learning programming logic.

To address these challenges, we introduce TiniScript, an intermediate programming language designed to simplify robotic programming for educational purposes while enabling real-time, wireless control. TiniScript’s minimalist syntax reduces instructions to a single line, such as F(2, 80), for commands like moving forward for 2 seconds at 80\% power, which allows users to bypass the need for complex coding. TiniScript bridges the gap between block-based programming and text-based coding by providing a user-friendly syntax that can be easily understood and implemented across a variety of robotic platforms. \cite{papert1993children, mitNews2016}

A significant innovation in TiniScript is its compatibility with any Bluetooth-enabled device—tablets, computers, or smartphones—allowing commands to be sent via a serial port directly to the robot. Equipped with a preloaded interpreter on the microcontroller, TiniScript eliminates the need for cable connections and repetitive code uploads. Commands can be transmitted wirelessly, which facilitates immediate responses from the robot, thus enhancing real-time interaction and iterative learning. This flexibility makes TiniScript adaptable across different robotic devices, creating a potential standard for educational robotics programming.

This paper explores TiniScript’s design philosophy, syntax, and applications, providing examples that illustrate its practical use in educational settings. Through its innovative approach, TiniScript aims to make robotic programming more accessible, empowering students to experiment and collaborate freely. By simplifying the coding process and enabling wireless control, TiniScript enhances the educational experience, fostering critical thinking and creativity among students as they engage with robotics in a dynamic and immersive manner.
\cite{cambridge2014handbook, mitMediaLab}.

\section{Design Philosophy of TiniScript}

TiniScript is designed as an intermediate programming language for educational robotics, aligning closely with the principles of STEM education. The primary goal behind TiniScript’s development is to create a simplified, yet powerful, programming environment that fosters interactive and iterative learning, essential in educational settings. By offering a minimalist syntax with single-line commands, TiniScript provides an accessible interface that allows students to command robots intuitively and focus on core programming concepts without being encumbered by hardware complexities.

The design of TiniScript emphasizes real-time interaction, which is critical for maintaining student engagement and promoting hands-on learning. Traditional programming methods require code to be burned repeatedly onto microcontrollers, a process that can be time-consuming and disrupt the flow of learning. TiniScript, in contrast, allows direct transmission of commands through a pre-installed interpreter on the microcontroller, which can interpret these instructions wirelessly via Bluetooth. This design drastically reduces the time needed to see the results of program changes, making it ideal for iterative learning where students can quickly test and refine their code.

By decoupling the programming environment from specific hardware intricacies, TiniScript enables educators and students to focus more on creative and cognitive aspects of robotics. This approach not only simplifies the setup process but also makes robotics programming more accessible to beginners, promoting inclusivity in educational robotics. Additionally, TiniScript’s flexibility in interfacing with a wide range of Bluetooth-enabled devices—such as tablets, smartphones, and computers—further enhances its adaptability and suitability for diverse educational contexts.

\begin{figure}[H]
    \centering
    \includegraphics[width=0.8\linewidth]{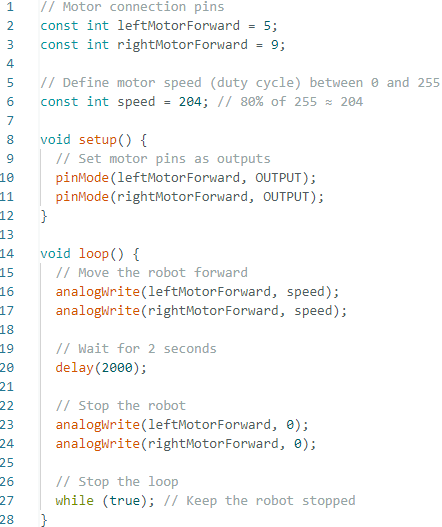}
    \caption{Arduino Code Example to move the robot forward for 2 seconds at 80\% power.}
    \label{fig:arduino_code}
\end{figure}

To illustrate this, consider the following example of code written in Arduino to make a robot move forward for 2 seconds at 80\% power, shown in Fig. \ref{fig:arduino_code}.

\begin{figure}[H]
    \centering
    \includegraphics[width=0.2\linewidth]{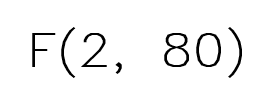}
    \caption{TiniScript Code Example to achieve the same movement with a simplified command.}
    \label{fig:tiniscript_code}
\end{figure}

In TiniScript, this same command can be executed with a single-line instruction, as shown in Fig. \ref{fig:tiniscript_code}.

\begin{figure}[H]
    \centering
    \includegraphics[width=0.8\linewidth]{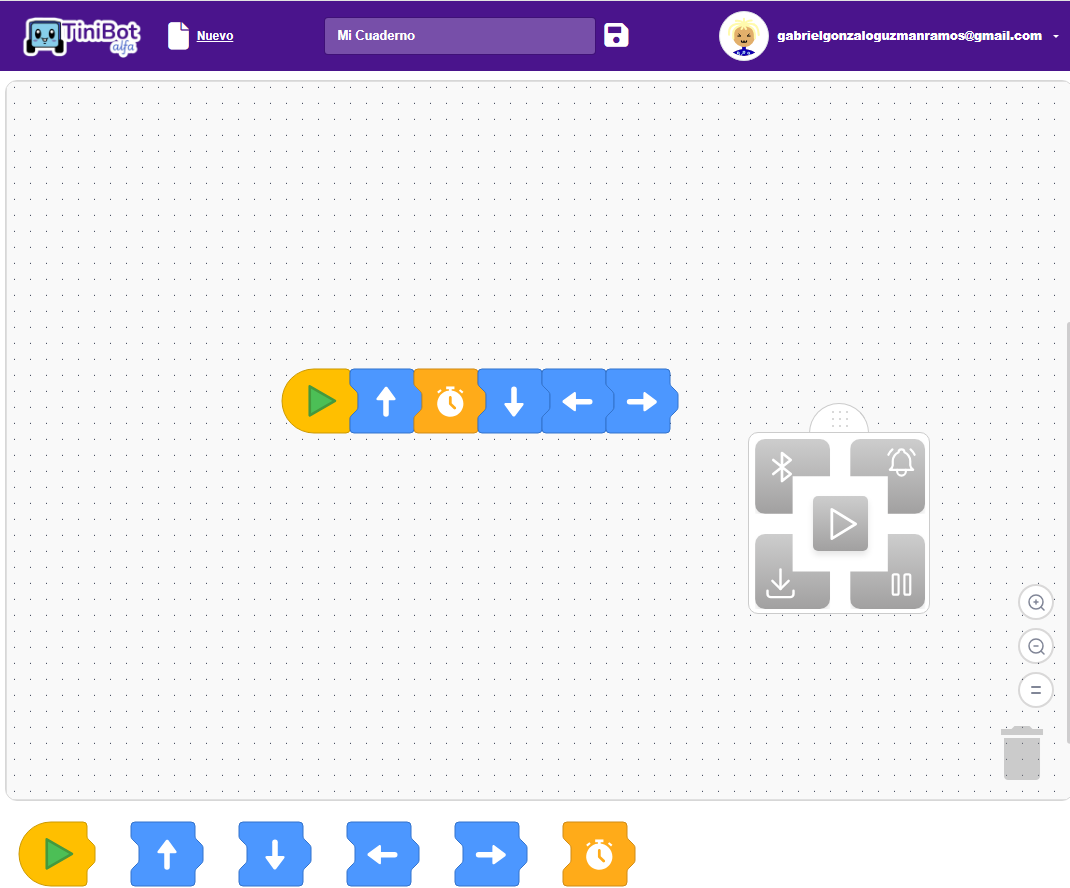}
    \caption{TiniBot interface for block-based programming, which converts blocks into TiniScript commands for easy use in educational robotics.}
    \label{fig:tinibot_interface}
\end{figure}

\begin{figure}[H]
    \centering
    \includegraphics[width=0.6\linewidth]{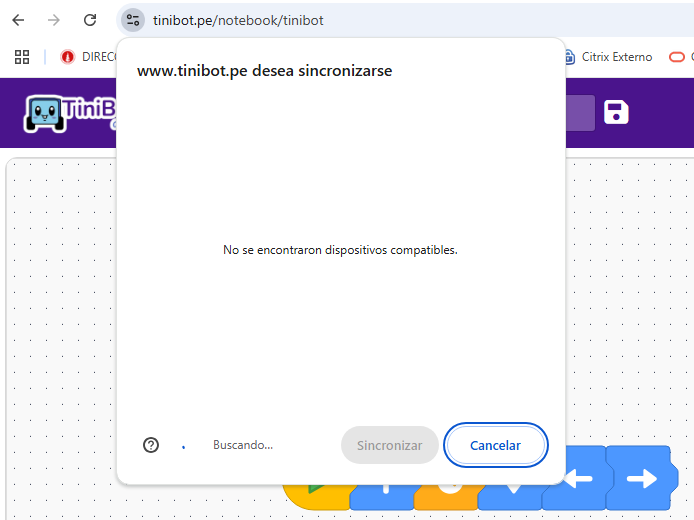}
    \caption{Synchronization process on TiniBot using Bluetooth to connect compatible devices.}
    \label{fig:tinibot_sync}
\end{figure}

One of the main advantages of TiniScript is its compatibility with a variety of devices, as it uses Bluetooth signals to transmit instructions. Any device with Bluetooth capabilities—such as tablets, smartphones, or computers—can send instructions to the robot, making TiniScript highly adaptable to diverse educational environments. To support this functionality, we have developed an online platform at \url{http://tinibot.pe}, where users can experiment with block-based programming that is automatically converted to TiniScript code, which is sent to the Arduino or ESP32. This site allows users to emulate Arduino and ESP32 code in TiniScript, making robotics programming accessible even to those unfamiliar with traditional code syntax. Fig. \ref{fig:tinibot_interface} shows the block-based programming interface available on the TiniBot platform, and Fig. \ref{fig:tinibot_sync} demonstrates the synchronization process via Bluetooth.

\section{Architecture of TiniScript}

\begin{figure}[H]
    \centering
    \includegraphics[width=0.9\linewidth]{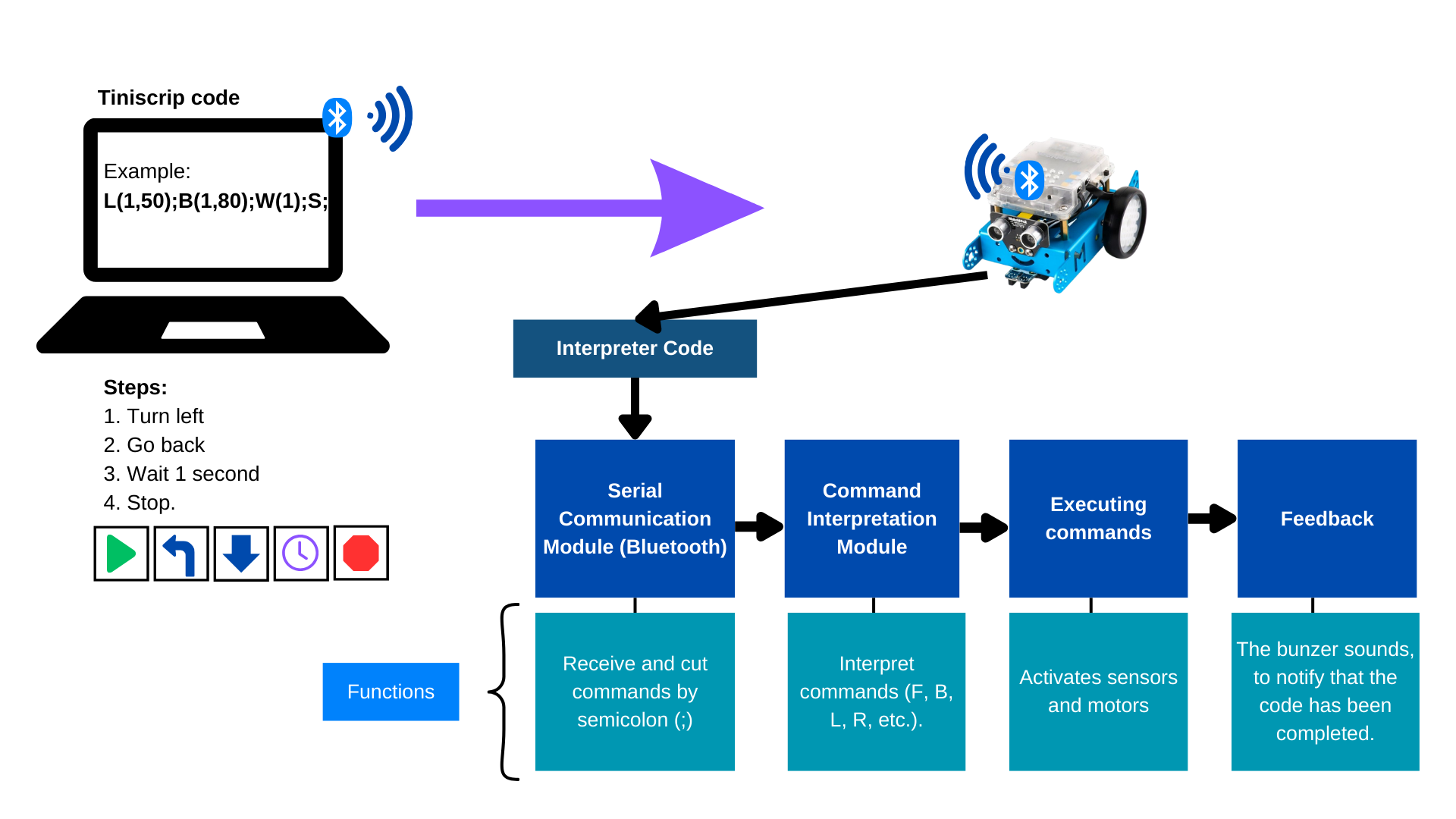}
    \caption{Architecture of TiniScript: command flow from the sending device to execution and feedback in the robot.}
    \label{fig:arquitectura_tiniscrip}
\end{figure}

The architecture of TiniScript is based on a serial communication system via Bluetooth, allowing the sending of commands from an external device (such as a computer, tablet, or smartphone) to the robot. This process of communication and execution of commands is managed by a code interpreter pre-installed on the robot's microcontroller, which interprets and executes the instructions in real time. Fig. \ref{fig:arquitectura_tiniscrip} illustrates the system architecture, showing the data flow from sending commands to the feedback provided to the user.

The functioning of TiniScript is structured in the following modules:\\

1. \textbf{Sending of commands}: The TiniScript commands, such as \textbf{L(1,50); B(1,80); W(1); S;}, are sent from a sending device via a Bluetooth connection. These commands are composed of a letter that indicates the action (for example, `F` to move forward, `B` to move backward) followed by a value that determines the duration or intensity of the action.\\

2. \textbf{Serial Communication Module}: This module receives the TiniScript commands and separates them by delimiters (for example, semicolon `;`), processing each command individually. This allows the robot to execute multiple actions in sequence without needing to reload the program onto the microcontroller.\\

3. \textbf{Command Interpretation Module}: Once the command is received and separated, the interpretation module translates it into specific instructions for the robot. The letters (`F`, `B`, `L`, `R`, `S`, etc.) represent actions such as moving forward, moving backward, turning left, turning right, and stopping, respectively. This module is responsible for identifying the indicated action and managing the associated parameters, such as speed or execution time.\\

4. \textbf{Command Execution}: The interpreted instructions are executed on the robot's hardware, directly controlling the motors and sensors to carry out the indicated actions. This module allows the robot to respond quickly and accurately to commands, following the sequence defined by the user in real time.\\

5. \textbf{Feedback}: Once the command sequence is completed, the system provides feedback to the user through a sound emitted by the robot's buzzer. This sound indicates that the program has finished, closing the interaction cycle and allowing the user to know that the robot has completed all assigned tasks.\\

\section{Syntax and Core Features of TiniScript}

TiniScript offers a simplified and efficient command structure, ideal for controlling robots in educational environments. The syntax of TiniScript is designed to be intuitive and easy to understand, allowing for the quick implementation of instructions through single-line commands. This section describes the basic structure of commands as well as the main features of TiniScript, including setup commands, movement instructions, sensor usage, and control structures.

\subsection{Syntax}
Each instruction in TiniScript follows the structure:

\begin{verbatim}
SETUP|INSTRUCTIONS
\end{verbatim}

where \texttt{SETUP} defines the start mode and \texttt{INSTRUCTIONS} contains the command to execute. Complex instructions can be composed using ``;'' as a separator.

\subsubsection{SETUP Commands}

The setup commands (\texttt{SETUP}) allow for controlling the start of instruction execution:

\begin{itemize}
    \item \texttt{PING|check\_connection}: Checks the connection with the robot.
    \begin{itemize}
        \item \textbf{Example:}
        \begin{verbatim}
        PING|check_connection
        \end{verbatim}
    \end{itemize}
    \item \texttt{SI|...}: Starts execution immediately.
    \begin{itemize}
        \item \textbf{Example:}
        \begin{verbatim}
        SI|F(5, 80)
        \end{verbatim}
    \end{itemize}
    \item \texttt{SB|...}: Starts execution when a button is pressed.
    \begin{itemize}
        \item \textbf{Example:}
        \begin{verbatim}
        SB|R(3, 60)
        \end{verbatim}
    \end{itemize}
\end{itemize}

\subsubsection{INSTRUCTIONS Commands}

\paragraph{Movement}
The movement commands allow controlling the robot's direction and speed:
\begin{itemize}
    \item \texttt{F(time, power)}: Moves forward for \texttt{time} seconds with a power of \texttt{power}.
    \item \texttt{B(time, power)}: Moves backward for \texttt{time} seconds with a power of \texttt{power}.
    \item \texttt{L(time, power)}: Turns left for \texttt{time} seconds with a power of \texttt{power}.
    \item \texttt{R(time, power)}: Turns right for \texttt{time} seconds with a power of \texttt{power}.
    \item \texttt{S}: Stops the robot.
    \item \texttt{W(seconds)}: Pauses the robot's execution for \texttt{seconds}.
\end{itemize}

\paragraph{Sensors}
Sensor measurements are stored in variables for use in conditional structures. The creation of new variables depends on the quantity and type of sensors in each kit. Basic sensors include:
\begin{itemize}
    \item \texttt{LIGHT\_R}: Stores the light value measured by the right sensor.
    \item \texttt{LIGHT\_L}: Stores the light value measured by the left sensor.
    \item \texttt{DISTANCE}: Stores the distance value measured by the sensor.
\end{itemize}

\paragraph{Comparators}
Comparators allow for creating conditions in conditional structures:
\begin{itemize}
    \item \texttt{=}: Equal to.
    \item \texttt{<}: Less than.
    \item \texttt{>}: Greater than.
    \item \texttt{<=}: Less than or equal to.
    \item \texttt{>=}: Greater than or equal to.
    \item \texttt{<>}: Not equal to.
\end{itemize}

\paragraph{Mathematical Operations}
TiniScript also allows performing mathematical operations within instructions:
\begin{itemize}
    \item \texttt{+}: Addition.
    \item \texttt{-}: Subtraction.
    \item \texttt{*}: Multiplication.
    \item \texttt{/}: Division.
    \item \texttt{\%}: Modulus.
    \item \texttt{ROUND(value, decimals)}: Rounds \texttt{value} to the specified number of \texttt{decimals}.
\end{itemize}

\paragraph{Boolean Values}
Boolean values and logical operators are used in conditional structures to control the program flow:
\begin{itemize}
    \item \texttt{TRUE}: Boolean true.
    \item \texttt{FALSE}: Boolean false.
    \item \texttt{AND}: Logical AND.
    \item \texttt{OR}: Logical OR.
    \item \texttt{NOT}: Logical NOT.
\end{itemize}

\paragraph{Conditionals}
Conditionals allow for executing instructions if a condition is met:
\begin{itemize}
    \item \texttt{IF(condition); INSTRUCTION; ENDIF}: Executes \texttt{INSTRUCTION} if \texttt{condition} is true.
    \begin{itemize}
        \item \textbf{Example:}
        \begin{verbatim}
        SI|IF(LIGHT_R > 100);F(4, 70);ENDIF
        \end{verbatim}
    \end{itemize}
\end{itemize}

\paragraph{Loops}
Loops allow for repeating an instruction a specified number of times:
\begin{itemize}
    \item \texttt{LOOP(n); INSTRUCTION; END\_LOOP}: Repeats \texttt{INSTRUCTION} \texttt{n} times.
    \begin{itemize}
        \item \textbf{Example:}
        \begin{verbatim}
        SI|LOOP(3);F(2, 50);END_LOOP
        \end{verbatim}
    \end{itemize}
\end{itemize}

\subsubsection{TiniScript Program Examples}

\begin{enumerate}
    \item \textbf{Start immediately and move forward for 5 seconds with power 80:}
    \begin{verbatim}
    SI|F(5, 80)
    \end{verbatim}
    
    \item \textbf{Start with the button and turn right for 3 seconds with power 60:}
    \begin{verbatim}
    SB|R(3, 60)
    \end{verbatim}
    
    \item \textbf{Move forward for 4 seconds with power 70 if the light measured by the right sensor is greater than 100:}
    \begin{verbatim}
    SI|IF(LIGHT_R > 100);F(4, 70);ENDIF
    \end{verbatim}
    
    \item \textbf{Ping to check the connection:}
    \begin{verbatim}
    PING|check_connection
    \end{verbatim}
    
    \item \textbf{Repeat 3 times: move forward for 2 seconds with power 50:}
    \begin{verbatim}
    SI|LOOP(3);F(2, 50);END_LOOP
    \end{verbatim}
\end{enumerate}

TiniScript is designed to be easy to learn and use, making robot programming accessible to students. Its simple and direct syntax allows students to focus on the logic and design of their programs rather than on the complex details of low-level programming.

\section{Comprehensive Example}
To illustrate the practical application and benefits of TiniScript, let’s consider a comprehensive example that encompasses various aspects of robotics programming, including movement, sensor interaction, conditional logic, and loops. This example will demonstrate how TiniScript can be used to create a program that enables a robot to navigate through an obstacle course, detect obstacles, and respond appropriately.

\subsection{Problem Statement}
Design a program in TiniScript for a robot to navigate through an obstacle course. The robot should move forward until it detects an obstacle within 10 cm. Upon detecting an obstacle, it should stop, turn right, and continue moving forward. This process should repeat continuously.

\subsection{Setup and Initial Instructions}
First, we need to configure the robot to start immediately upon receiving the instructions:

\begin{lstlisting}[style=mystyle]
SI|START
\end{lstlisting}

\subsection{Movement and Obstacle Detection}
Next, we define the instructions for the robot to move forward and continuously check for obstacles using a distance sensor. The robot will employ a loop to constantly check the distance to obstacles and respond accordingly.

\begin{lstlisting}[style=mystyle]
SI|LOOP(FOREVER);
  F(1, 80);
  DISTANCE;
  IF(DISTANCE < 10);
    STOP;
    R(1, 60);
  ENDIF;
END_LOOP
\end{lstlisting}

\subsection{Detailed Explanation}
\begin{itemize}
  \item \texttt{SI|START}: Configures the robot to start immediately.
  \item \texttt{SI|LOOP(FOREVER);}: Initiates an infinite loop, allowing the robot to continuously execute the following instructions.
  \item \texttt{F(1, 80);}: Moves the robot forward for 1 second with a power of 80.
  \item \texttt{DISTANCE;}: Reads the distance to the obstacle and stores the value in the \texttt{DISTANCE} variable.
  \item \texttt{IF(DISTANCE < 10);}: Checks if the distance to the obstacle is less than 10 cm.
  \item \texttt{STOP;}: Stops the robot if the condition is met.
  \item \texttt{R(1, 60);}: Turns the robot right for 1 second with a power of 60 to avoid the obstacle.
  \item \texttt{ENDIF;}: Ends the conditional statement.
  \item \texttt{END\_LOOP;}: Ends the infinite loop, although in this case the loop is set to run indefinitely, so it will restart.
\end{itemize}

\subsection{Wireless Communication}
This example program can be sent to the robot wirelessly via Bluetooth or WiFi. The TiniScript interpreter, preloaded on the robot’s microcontroller, will interpret these instructions and execute them, allowing for real-time interaction and iterative learning.

\begin{lstlisting}[style=mystyle]
SI|START;LOOP(FOREVER);F(1, 80);DISTANCE;IF(DISTANCE < 10);STOP;R(1, 60);ENDIF;END_LOOP
\end{lstlisting}

\subsection{Advantages of TiniScript in This Scenario}
The use of TiniScript in this task offers several advantages:

\begin{itemize}
  \item \textbf{Simplified Syntax:} The single-line instruction syntax makes the program easy to write and understand, ideal for beginners.
  \item \textbf{Real-Time Interaction:} Instructions can be sent wirelessly, allowing for immediate feedback and quick adjustments.
  \item \textbf{Iterative Learning:} Students can quickly test and modify their programs, fostering a deep understanding of robotics concepts.
  \item \textbf{Accessibility:} The simplicity and efficiency of TiniScript make it accessible to beginners and suitable for educational environments, where learning and experimentation are key.
\end{itemize}

\section{Conclusion}

TiniScript has established itself as a valuable tool in the field of educational robotics, aligning closely with the principles of STEM education. Its minimalist single-line instruction syntax allows students and educators to simplify the programming process, facilitating a smoother and more effective transition from conceptual learning to practical application. This intermediate programming language not only significantly reduces the program loading times on microcontrollers but also facilitates real-time interaction and iterative learning through wireless communication.\\

One of the key advantages of TiniScript is its ability to send instructions directly to the robot via Bluetooth, eliminating the need for cable connections and repeated code uploads. This feature, combined with a pre-installed interpreter on the microcontroller, allows students to focus more on the logic and creativity of their projects rather than on the technical complexities of the hardware. By decoupling the programming environment from the specifics of the hardware, TiniScript empowers educators and students to explore and develop their robotics skills in a more accessible and engaging way.\\

Additionally, the ability to modify and test code wirelessly enhances collaborative learning environments, where multiple students can interact with and program the robot simultaneously. TiniScript’s compatibility with a variety of devices and its modular architecture make it highly adaptable to diverse educational contexts. The integration with block-based programming platforms, such as the online interface available at \url{http://tinibot.pe}, further enhances its accessibility by allowing users to seamlessly convert block-based code to TiniScript commands.\\

Through its innovative design and practical approach, TiniScript bridges the gap between beginner-friendly block-based programming environments and advanced microcontroller programming. It supports the development of logical thinking, creativity, and problem-solving skills, fundamental objectives in STEM learning. By making robotics programming more accessible and effective in developing critical thinking skills, TiniScript has the potential to transform the way educational robotics is taught and learned, fostering a new generation of innovators and thinkers in the field of robotics.

\end{document}